\documentclass{article}
\usepackage{arxiv}

\usepackage{lmodern}
\usepackage[style=base,figurename=Fig.,labelfont=bf,labelsep=period]{caption}
\usepackage{subcaption}
\usepackage{amsmath}
\usepackage{newtxtext,newtxmath}
\usepackage[colorlinks=true,citecolor=blue,linkcolor=blue,urlcolor=blue]{hyperref}

\usepackage[utf8]{inputenc} % allow utf-8 input
\usepackage[T1]{fontenc}    % use 8-bit T1 fonts
\usepackage{booktabs}       % professional-quality tables
\usepackage{nicefrac}       % compact symbols for 1/2, etc.
\usepackage{microtype}      % microtypography
\usepackage{graphicx}
\usepackage{natbib}
\usepackage{doi}

\usepackage{algorithm}
\usepackage{algpseudocode}
\usepackage{enumitem}
\title{	Interpretable Deep Reinforcement Learning for Element-level Bridge Life-cycle Optimization }
\date{} 					% Or removing it

\author{ \href{https://orcid.org/0000-0003-1808-9050}{\includegraphics[scale=0.06]{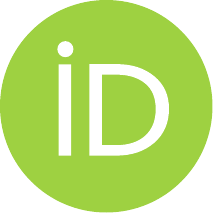}\hspace{1mm}Seyyed Amirhossein~Moayyedi}\thanks{Ph.D. Candidate} \\
	Department of Civil and Environmental Engineering\\
	Portland State University\\
	Portland, OR 97201 \\
	\texttt{amirm@pdx.edu} \\
	\And
	\href{https://orcid.org/0000-0003-0959-6333}{\includegraphics[scale=0.06]{orcid.pdf}\hspace{1mm}David Y.~Yang} \thanks{Assistant Professor, Corresponding Author}\\
	Department of Civil and Environmental Engineering\\
	Portland State University\\
	Portland, OR 97201 \\
	\texttt{david.yang@pdx.edu} \\
}

\renewcommand{\headeright}{Moayyedi and Yang}
\renewcommand{\undertitle}{}
\renewcommand{\shorttitle}{Interpretable RL for Life-cycle Optimization}

\hypersetup{
pdftitle={Interpretable Deep Reinforcement Learning for Element-level Bridge Life-cycle Optimization},
pdfsubject={cs.AI, cs.LG},
pdfauthor={Seyyed Amirhossein~Moayyedi, David Y.~Yang},
pdfkeywords={Interpretable machine learning, Reinforcement learning, Life-cycle optimization, Bridge management},
}
\begin{document}

\maketitle

% Please include an abstract:
\begin{abstract}
The new Specifications for the National Bridge Inventory (SNBI), in effect from 2022, emphasize the use of element-level condition states (CS) for risk-based bridge management. Instead of a general component rating, element-level condition data use an array of relative CS quantities (i.e., CS proportions) to represent the condition of a bridge. Although this greatly increases the granularity of bridge condition data, it introduces challenges to set up optimal life-cycle policies due to the expanded state space from one single categorical integer to four-dimensional probability arrays. This study proposes a new interpretable reinforcement learning (RL) approach to seek optimal life-cycle policies based on element-level state representations. Compared to existing RL methods, the proposed algorithm yields life-cycle policies in the form of oblique decision trees with reasonable amounts of nodes and depth, making them directly understandable and auditable by humans and easily implementable into current bridge management systems. To achieve near-optimal policies, the proposed approach introduces three major improvements to existing RL methods: (a) the use of differentiable soft tree models as actor function approximators, (b) a temperature annealing process during training, and (c) regularization paired with pruning rules to limit policy complexity. Collectively, these improvements can yield interpretable life-cycle policies in the form of deterministic oblique decision trees. The benefits and trade-offs from these techniques are demonstrated in both supervised and reinforcement learning settings. The resulting framework is illustrated in a life-cycle optimization problem for steel girder bridges.
\end{abstract}


\section{Introduction}
Bridges deteriorate over their service life. Therefore, various interventions such as maintenance, repair, and rehabilitation are required over the structural life-cycle to ensure that the structures can always operate with a sufficient safety margin. Life-cycle optimization refers to the process of optimizing the timing and the type of life-cycle actions over the course of a bridge’s remaining service life. The output of this optimization is a life-cycle policy consisting of a collection of action triggers \citep{fhwa_bridge_2020}.

Each action trigger recommends a life-cycle action based on prescribed rules related to bridge condition and sometimes the remaining service life \citep{fhwa_bridge_2020}. For instance, one common action trigger is to rehabilitate the component if its general condition rating (GCR) is at or below four \citep{fhwa_bridge_2018}. The goal of life-cycle optimization is to minimize the total life-cycle cost that encompasses the costs of all life-cycle actions as well as the gradually increasing risks of unexpected service disruptions due to deterioration and a lack of timely actions \citep{yang_framework_2026}. This risk-based approach corresponds to a sequential decision-making problem, where in each year a decision maker needs to balance the immediate cost-benefit of a life-cycle action and the decision's implications for future deterioration and costs.

The U.S. Specifications for the National Bridge Inventory (SNBI), which went into effect in 2022 \citep{fhwa_specifications_2022}, have brought about new challenges to the current practice of bridge life-cycle optimization. A major shift in the 2022 SNBI is the representation of bridge condition, which transitions from a general condition rating (GCR) for each of the three major components (deck, superstructure and substructure) to element-level condition state (CS) distributions for over 80 national bridge elements (NBE) defined in the AASHTO Manual \citep{AASHTO2019} and many more agency-defined elements. While this increases the granularity of bridge condition data, it significantly expands the state space upon which life-cycle decisions are made. Even for a single element, this transition means its condition is now represented by a vector of four probabilities (representing the proportions of four CSs) instead of a single categorical integer ranging from GCR~=~0 to GCR~=~9.

Currently, life-cycle policies using this element-level condition data are commonly implemented in practice through either (a) translators that revert element CS data back to GCRs \citep{bektas_using_2013}, allowing adoption of previous GCR-based policies, or (b) action triggers based on a bridge health index (BHI), i.e., an empirical function that aggregates element CS distributions into a single indicator for decision-making \citep{chase_synthesis_2016}. For instance, Caltrans developed the California BHI and uses it to flag bridges for major actions if the BHI falls below 70 \citep{shepard_california_2001}. The primary drawback of this current practice is that these aggregated life-cycle policies do not fully capitalize on the finer resolution of bridge condition afforded by the element-level data. Consequently, the derived life-cycle policies may not be optimal, or even near-optimal, for reducing the risk-included total life-cycle cost.

To address this expanded state space for decision-making, deep reinforcement learning (DRL) methods have been increasingly leveraged to optimize life-cycle policies. Andriotis and Papakonstantinou first established the algorithm and application of DRL for the life-cycle management of deteriorating structures, both without \citep{Andriotis2019} and with \citep{Andriotis2021} constraints. Following their seminal work, subsequent studies have developed DRL methods for life-cycle optimization based on structural reliability \citep{Yang2022}, load rating factor \citep{cheng_decision-making_2021}, PennDOT-specific GCRs \citep{bhattacharya_district-level_2025}, and GCRs augmented by bridge parameters \citep{ghavidel_risk-based_2024}. However, these DRL-based life-cycle policies are encoded within deep neural network models, which are notoriously unexplainable regarding their specific model outputs (i.e., life-cycle actions in this context). As a result, current DRL methods yield life-cycle policies that cannot be easily understood, inspected, adjusted, or audited by human decision-makers. This ``black-box'' nature of DRL solutions significantly hinders their practical application and adoption in the management of critical infrastructure assets such as bridges.

Beyond bridge life-cycle optimization and transportation asset management, a growing body of literature focuses on the explainability and interpretability of machine learning models \citep{sahin_unlocking_2025}. Strictly speaking, explainability refers to the ability to gain \textit{ex post} insights that summarize the reasoning behind model predictions. In contrast, interpretability describes the intrinsic characteristics of a prediction or decision-making model that allows it to be directly understood by humans with minimal postprocessing \citep{naser_machine_2023}. Commonly referenced interpretable models include linear or logistic regression models, as well as tree-based models \citep{naser_machine_2023}. The latter is particularly popular for decision-making tasks because it most closely resembles human cognitive processes in critical fields such as medicine and healthcare \citep{marewski_heuristic_2012,phillips_fftrees_2017}.

Due to the widespread adoption of neural networks in engineering (and civil engineering is no exception), the focus of explainable and interpretable artificial intelligence within the field has skewed toward explainability rather than intrinsic interpretability. For instance, \cite{zaker_esteghamati_developing_2021} investigated explainable, data-driven surrogate models for performance-based seismic design. In the context of life-cycle optimization, \cite{yang_deep_2022} demonstrated that neural network-based life-cycle policies can be explained via policy visualization, advocating for the use of imitation learning to extract interpretable proxies. Empirical results from that study also showed that interpretable tree-based policies, summarized from policy visualization, can be made understandable by humans while simultaneously functioning as effectively as more complex, neural network-based policies.

Despite this demonstrated potential, inherently interpretable models remain underexplored in bridge life-cycle optimization due to two major hurdles: (a) the ineffectiveness of linear or logistic models in handling complex sequential decision-making, and (b) the lack of gradients in tree-based models, which prevents their use in gradient-based RL algorithms. To fill in this gap, the present study proposes an interpretable reinforcement learning approach that directly derives tree-based policies for element-level bridge life-cycle optimization. This novel approach utilizes differentiable soft tree models for policy approximation, alongside annealing, regularization and pruning schemes designed to convert the soft tree model into a simplified interpretable oblique decision tree. The following sections detail this methodology, validate it via supervised learning, and present an application case study focusing on element-level bridge life-cycle management.

\section{Methodology}

\subsection{Oblique Decision Tree as an Interpretable Model}
A standard decision tree compares a single feature against a threshold value at each internal node, typically splitting the prediction into two alternative branches (i.e., binary splits). Although multiway splits are occasionally used, they can also be achieved through a series of binary splits \citep{Hastie2017}. Therefore, they are not further considered in this study. A complete decision path is formed by a chain of binary splits that leads to a deterministic label for either regression or classification tasks. Because our subsequent analysis focuses on discrete decisions, the discussion hereafter centers on classification tasks.  While easily interpretable and widely used in human decision-making, standard decision trees struggle to handle decision boundaries that are not well aligned with the input features. When applied to problems with such non-orthogonal decision boundaries, standard decision trees often produce overly deep models that are less interpretable and more prone to overfitting \citep{Geron2019}.

To overcome the aforementioned limitations of standard decision trees, linear combinations of input features can be used at internal nodes for binary splits. In particular, the binary split at an internal node can be expressed as:
\begin{equation}
    \begin{split}
        \mathbf{w}_j^\intercal \mathbf{x} + b_j &\leq 0 \quad \text{for left branch} \\
        \mathbf{w}_j^\intercal \mathbf{x} + b_j &> 0 \quad \text{for right branch}
    \end{split}
    \label{eq:oblique_split}
\end{equation}
where $\mathbf{x}$ = input features; $\mathbf{w}_j$ = feature weights of $j$-th internal node; $b_j$ = bias of $j$-th internal node, which serves functionally as the decision threshold. Geometrically, each split defines a hyperplane in the feature space, allowing the tree to form oblique decision boundaries that reflect interactions among the input features. This approach gives rise to what are known as oblique decision tree models \citep{Hastie2017}.

\begin{figure}
    \centering
    \includegraphics[width=5.875in]{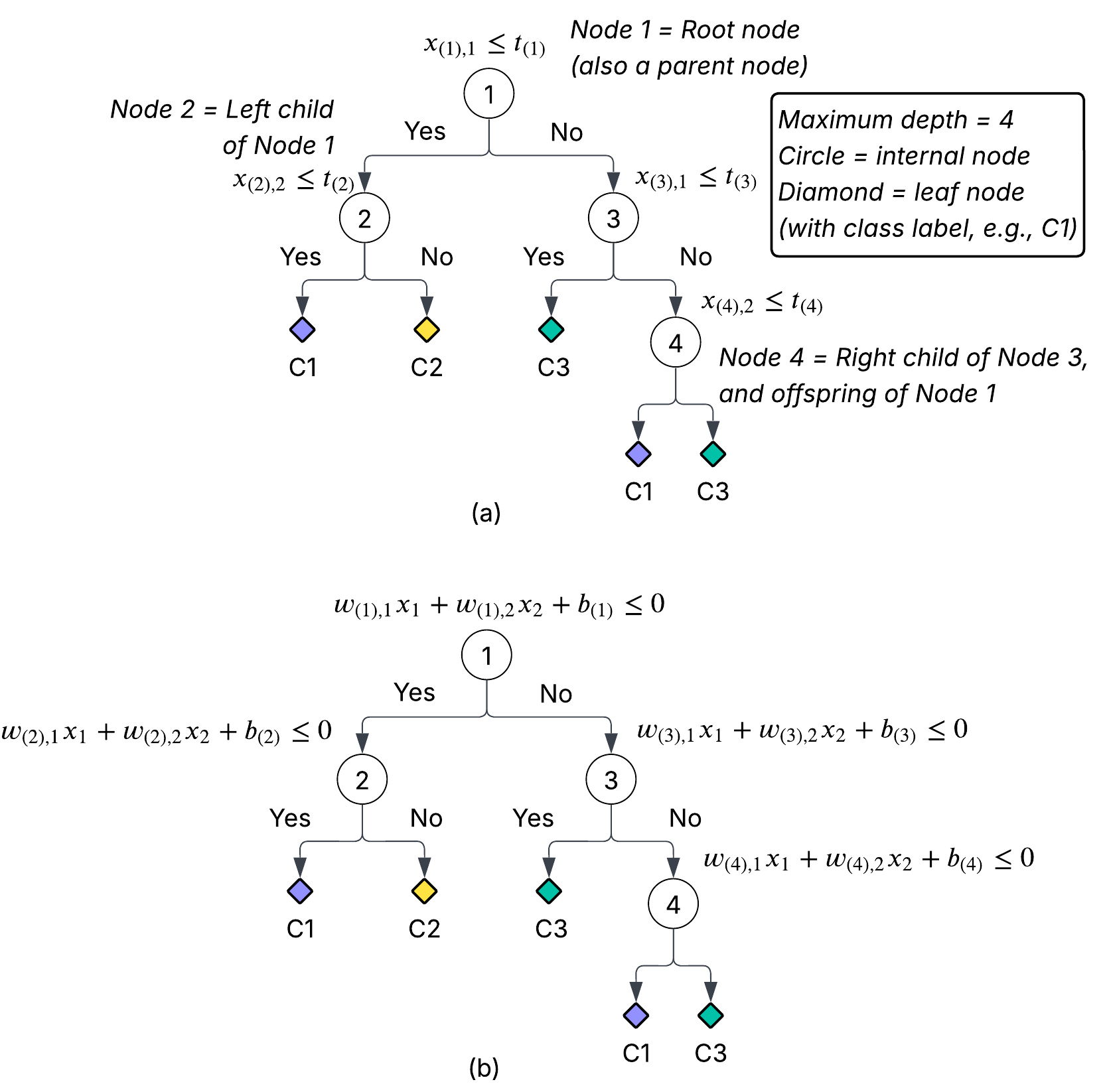}
    \caption{Tree-based interpretable decision models: (a) standard decision tree and (b) oblique decision tree.}
    \label{fig:tree_models}
\end{figure}

Fig.~\ref{fig:tree_models} illustrates two decision models using standard and oblique decision trees, respectively, based on two input features. Relevant terminology used throughout this study is also introduced within the figure. As illustrated, when the tree models are shallow, both models remain easily  inspectable and interpretable. Additionally, standard decision trees can be viewed as a special case of oblique decision trees, in which only one of the feature weights is non-zero. In the context of element-based bridge life-cycle policies, the input features are the CS proportions of key bridge elements. Therefore, the linear combination of input features used in oblique decision trees is strongly associated with the formulation of a BHI, and the binary split is akin to an action trigger used to define a life-cycle policy.

Despite this close association with element-level action triggers, the discreteness inherent to binary splits prevents the use of gradient-based optimization techniques for determining the decision thresholds or feature weights in oblique decision trees \citep{Hastie2017}. This challenge effectively precludes their direct adoption into RL algorithms for optimal decision-making.

\subsection{Soft Tree and the Corresponding Frozen Tree}
To overcome the computational challenges mentioned above, differentiable tree models have been developed by replacing conventional binary splits with probabilistic gating functions and substituting the deterministic predictions at the leaf nodes with class probabilities or logits conditioned on arriving at a given leaf node. This approach transitions from ``hard'' deterministic decisions to ``soft'' class probabilities. This variant of tree models is referred to in the literature as soft trees \citep{frosst_distilling_2017} or hierarchical mixtures of experts \citep{Hastie2017}. For consistency with standard and oblique decision trees, the term ``soft trees'' is preferred in the following discussion.

\begin{figure}
    \centering
    \includegraphics[width=1.792in]{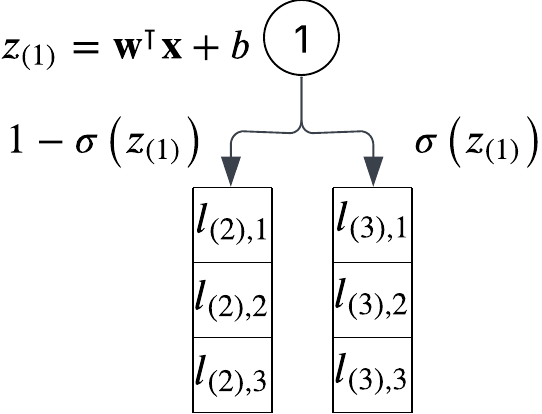}
    \caption{A simple soft tree with one internal node for three-class classification.}
    \label{fig:simple_soft_tree}
\end{figure}

Fig.~\ref{fig:simple_soft_tree} shows the most simplistic form of a soft tree model, which involves only one internal node and three classes for a classification task. The sigmoid function is employed as the gating function, assigning probabilities for entering the two branches originating from the internal node. For each child leaf node, three logits are specified to compute the conditional class probabilities. According to the law of total probability, the unconditional probability of a specific class $k$, given the input features $\mathbf{x}$, can be expressed as follows:
\begin{equation}
    p_k(\mathbf{x}) = \left[1 - \sigma\left( \mathbf{w}^\intercal\mathbf{x}+b \right) \right] \cdot \frac{\exp{\left[ l_{(2),k} \right]}}{\sum_{j=1}^K{\exp{\left[ l_{(2),j} \right]}} } + \sigma\left( \mathbf{w}^\intercal\mathbf{x}+b \right) \cdot \frac{\exp{\left[ l_{(3),k} \right]}}{\sum_{j=1}^K{\exp{\left[ l_{(3),j} \right]}} }
    \label{eq:simple_soft_tree}
\end{equation}
where $p_k(\mathbf{x})$ = probability of class $k$; $\sigma(\cdot)$ = sigmoid function defined later; $\mathbf{w}$ and $b$ = feature weights and bias, respectively, of the internal node (node number is dropped since we only have one internal node); $l_{(2),k}$ and $l_{(2),j}$ = logits of classes $k$ and $j$, respectively, at Node 2 (leaf node); $l_{(3),k}$ and $l_{(3),j}$ = counterparts at Node 3 (leaf node); $K$ = total number of classes ($K=3$).

\begin{figure}
    \centering
    \includegraphics[width=3.25in]{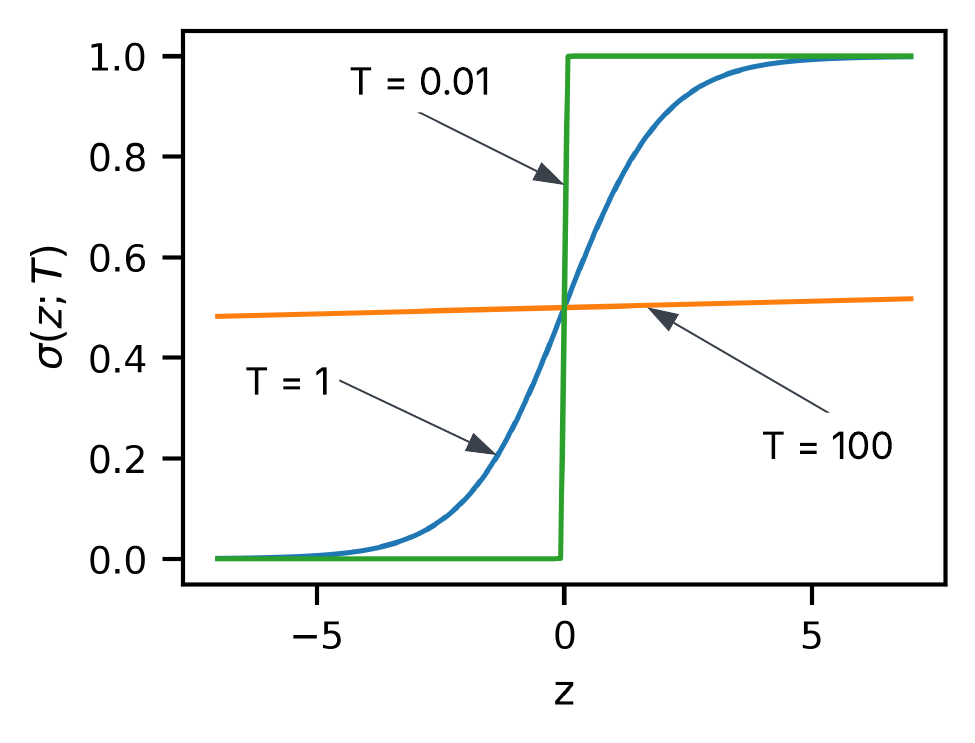}
    \caption{Temperature-controlled sigmoid function.}
    \label{fig:sigmoid}
\end{figure}

Instead of relying on the standard sigmoid function, we leverage a so-called temperature-controlled sigmoid function given as follows:
\begin{equation}
    \sigma(z;T) = \frac{1}{1+\exp{\left( -z/T \right)}}
    \label{eq:temp_sigmoid}
\end{equation}
where $z$ = output of an internal node, e.g., $z=\mathbf{w}^\intercal\mathbf{x}+b$ in the previous example; $T$ = temperature, with $T=1$ corresponding to the standard sigmoid function. As the temperature decreases, the sigmoid function gradually approaches the unit step function, as shown in Fig.~\ref{fig:sigmoid}. This asymptotic behavior enables the conversion of a soft tree model into an oblique decision tree. Specifically, if a soft tree uses a sigmoid gating function with a relatively low temperature (e.g., $T=0.01$), the corresponding oblique decision tree is obtained through the following steps:
\begin{itemize}
    \item At each internal node, replace the sigmoid gating function with a binary split using the same feature weights and bias.
    \item At each leaf node, replace the class logits (or probabilities) with a deterministic label based on the most likely class.
\end{itemize}
We refer to this process as ``freezing'' the soft tree model, and the resulting oblique decision tree is called a ``frozen'' tree model, which shares the same depth and internal node number as the original soft tree model.

Although soft trees provide differentiable class probabilities and the procedure above offers a way to obtain oblique decision trees, two obstacles still remain to achieving interpretable decision models that can be incorporated into RL algorithms. First, sigmoid functions at low temperatures present numerical challenges for gradient-based optimization algorithms due to their resemblance to a non-differentiable, discrete function. Second, the soft and the corresponding frozen trees are always fully grown---meaning every internal node produces exactly two child nodes---causing the total number of internal nodes to grow exponentially with tree depth. The increased complexity compromises model interpretability. The subsequent sections detail measures to overcome these obstacles.

\subsection{Temperature Annealing of Gating Function}
To support both the gradient-based optimization and the subsequent conversion to oblique decision trees, a temperature annealing schedule is proposed during the training of a soft tree model. Specifically, the temperature parameter $T$ in Eq.~\ref{eq:temp_sigmoid} is gradually decreased according to a predefined schedule. This procedure enables the model to begin training with well-defined gradients and progressively transition toward sharper, near-deterministic decisions, from which the final oblique decision tree is extracted. For both supervised and reinforcement learning tasks, the following annealing schedule is implemented:
\begin{itemize}
    \item Initialize $T_0$: Training can start with standard sigmoid function (e.g., $T_0=1$), ensuring well-defined gradients and stable optimization during the early stages of gradient-based training.
    \item Anneal $T$: The temperature is gradually decreased following an exponential schedule:
    \begin{equation}
        T(s) = T_0 \cdot \left( \frac{T_{\min}}{T_0} \right)^{s/S}
        \label{eq:annealing}
    \end{equation}
    where $s$ = stage of the annealing schedule (starting from 0), which may include several gradient descent steps; $S$ = total number of stages during the training process; $T_{\min}$ = minimum temperature at the end of the training.
\end{itemize}

\subsection{Regularization and Pruning of Soft Tree Models}

The complexity of a soft tree and its corresponding frozen tree arises from (a) the excessive number of internal nodes as the depth increases, and (b) the complexity of the decision rules at each internal node. To control model complexity and achieve interpretability, different regularization schemes can be incorporated during the training of a soft tree model. Specifically, $L_1$ regularization is leveraged in this study due to its potential to automatically filter irrelevant input features \citep{Bishop2006}, driving sparse models with a small number of internal nodes and simple decision rules at each node. Mathematically, $L_1$ regularization penalty can be expressed as follows:
\begin{equation}
    L_1 (\mathbf{W}) = \sum_{i \in \mathcal{N}} {\|\mathbf{w}_{(i)}\|_1}
    \label{eq:l1_reg}
\end{equation}
where $\mathbf{W}$ = matrix containing all feature weights of internal nodes; $\|\cdot\|_1$ = Manhattan norm of a vector. Note that only the feature weights are included in the regularization penalty. The biases are excluded due to their important role as decision thresholds in the binary splits. The penalty in Eq.~\ref{eq:l1_reg} is scaled and added to the loss function in supervised or reinforcement learning problems. 

A soft tree model trained with regularization can be similarly converted into an oblique decision tree using the same freezing process. Due to regularization-induced diminishing weights, the corresponding frozen tree can be significantly simplified. At each internal node, a sparse set of feature weights ensures that only the most relevant features are involved in the decision-making process. In instances where only one relevant feature remains, the oblique tree node reduces to an internal node typical of a standard decision tree. In the most extreme cases where all feature weights are diminished at an internal node, the decision at the given internal node becomes trivial and is determined solely by the sign of the node bias. In this scenario, a trivial internal node (defined by all close-to-zero feature weights) can be pruned from the oblique decision tree. Specifically, the trivial node is pruned by directly connecting its first non-trivial ancestor node to the first non-trivial offspring node (or a leaf node). As a result, the frozen tree is no longer a fully grown tree, thereby simplifying the decision-making process. Fig.~\ref{fig:prune_trivial} provides a recursive pseudo code to implement the process described previously. 

\begin{figure}
    \centering
    \fbox{
    \begin{minipage}{0.95\textwidth}
        \small
        \input{figures/prune_trivial}
        \normalsize
    \end{minipage}
    }
    \caption{Pseudo code for pruning trivial nodes}
    \label{fig:prune_trivial}
\end{figure}

Besides removing trivial nodes, the frozen tree can be further simplified by pruning nodes along infeasible decision paths caused by logically contradictory conditions. For instance, assume an internal node is reached because its parent node enforces the condition $x_1 + x_2 \leq 0$. If at the same time a previous ancestor node has already enforced $x_1 + x_2 > 5$, the internal node and all of its offspring nodes (i.e., subtree from the internal node) can never be reached, allowing the subtree to be pruned without ill effects.

\begin{figure}
    \centering
    \includegraphics[width=5.753in]{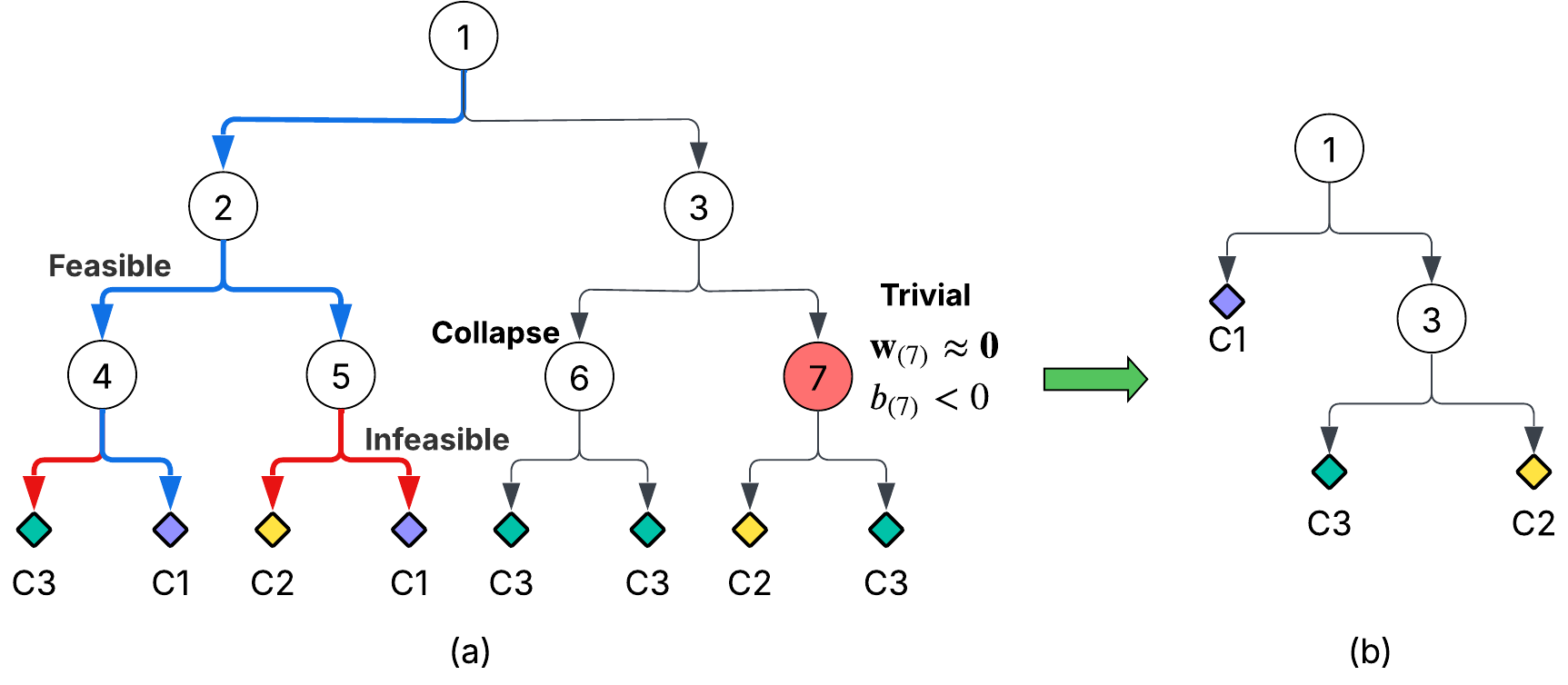}
    \caption{Example of pruning frozen tree: (a) original frozen tree and (b) oblique decision tree after pruning.}
    \label{fig:pruning_example}
\end{figure}

This process is implemented by analyzing the feasibility of a supplementary linear programming (LP) problem. Consider, for example, the 4-depth frozen tree shown in Fig.~\ref{fig:pruning_example}. For the decision path from Node 1 to Node 5, the weights and biases along the decision path are collected into the following matrix and vector:
\begin{equation}
    \mathbf{A} \gets \left[ \mathbf{w}_{(1)}, \mathbf{w}_{(2)}, \mathbf{w}_{(5)} \right] \quad \text{and} \quad \mathbf{d} \gets \left[ -b_{(1)}, -b_{(2)}, -b_{(5)} \right] ^\intercal
    \label{eq:LP_A_and_b}
\end{equation}
To check the feasibility of reaching the left child of Node 5, the following LP problem is formulated and analyzed:
\begin{equation}
    \min{\mathbf{c}^\intercal \mathbf{x}} \quad \text{subject to} \quad \mathbf{A} \mathbf{x} \leq \mathbf{d}
    \label{eq:LP_left}
\end{equation}
where $\mathbf{c}$ = a vector of non-zero constants to define a linear objective function; since the focus herein is to determine feasibility, $\mathbf{c}$ can be set arbitrarily. If no feasible set is found, the left branch of Node 5 can be pruned. To check the feasibility of reaching the right child of Node 5, one can simply flip the inequality sign for the constraint in Eq.~\ref{eq:LP_left}. Fig.~\ref{fig:prune_infeasible} presents the pseudo code to recursively implement this process within the frozen tree.

\begin{figure}
    \centering
    \fbox{
    \begin{minipage}{0.95\textwidth}
        \small
        \input{figures/prune_infeasible}
        \normalsize
    \end{minipage}
    }
    \caption{Pseudo code for pruning infeasible paths}
    \label{fig:prune_infeasible}
\end{figure}

A third routine to simplify the frozen tree involves collapsing internal nodes that possess two leaf child nodes with identical labels, as demonstrated by Node~6 in Fig.~\ref{fig:pruning_example}. In this case, the internal node is simply converted into a single leaf node. Using the three routines described above, the frozen tree depicted in Fig.~\ref{fig:pruning_example}(a) can be simplified to the oblique decision tree in Fig.~\ref{fig:pruning_example}(b). Note that the execution sequence of these three routines matters: it is best to remove trivial nodes first, followed by pruning infeasible paths, and ending with the collapse of identical leaf child nodes.

\subsection{Tree-based Interpretable Deep Learning Framework}
\begin{figure}
    \centering
    \includegraphics[width=6.169in]{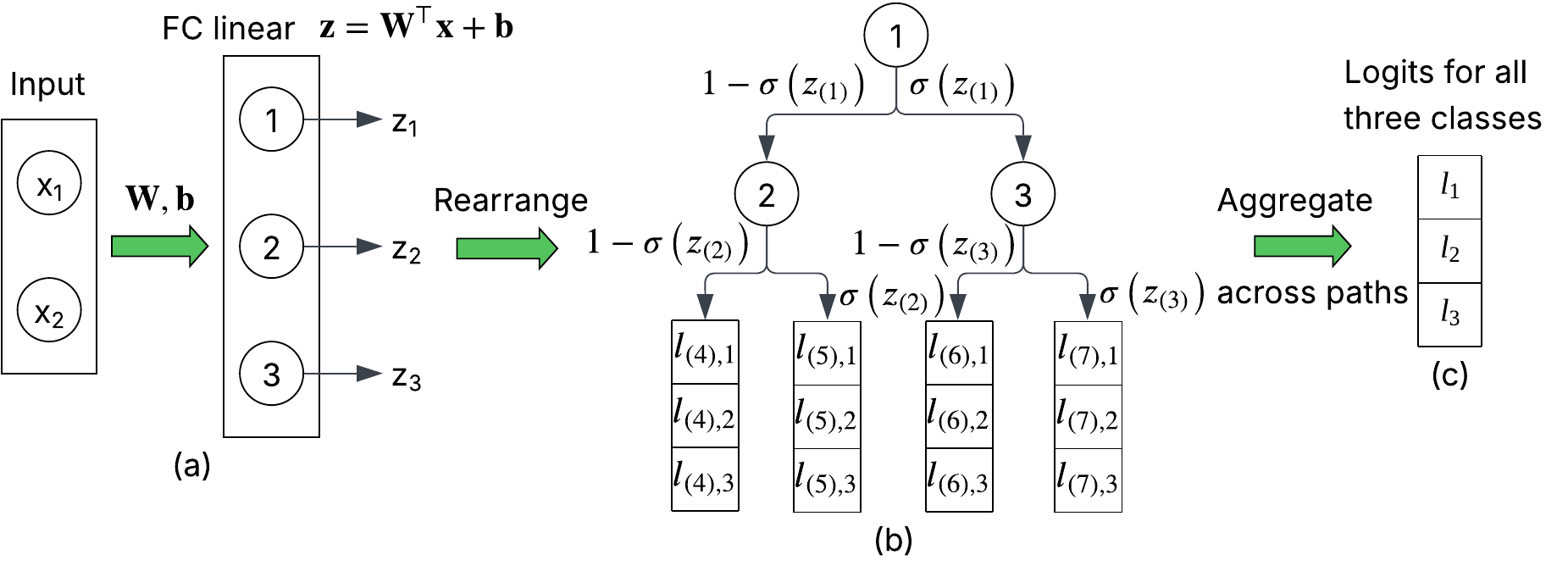}
    \caption{Implementation of a differentiable soft tree model (depth = 3): (a) neural network layers, (b) rearrangement to soft tree structure, and (c) output of class logits.}
    \label{fig:softtree_implement}
\end{figure}

The soft tree models described previously can be implemented within a neural network architecture, followed by custom differentiable operations to yield the logits for different classes (in classification tasks) or actions (in RL tasks). As an example, Fig.~\ref{fig:softtree_implement} illustrates this process using a simple soft tree model with two input features, three internal nodes, and three output classes. In Fig.~\ref{fig:softtree_implement}(a), the process begins with a single, fully connected linear neural network layer containing the same number of neurons as the total number of internal nodes. The weights and bias of each neuron correspond directly to the weights and bias of an internal node. In Fig.~\ref{fig:softtree_implement}(b), the outputs from the linear layer are converted to branch probabilities by rearranging them based on the soft tree structure and the sigmoid gating function. The class logit at each leaf node is then computed using the conditional logits stored at that specific leaf node and the product of all probabilities along the decision path leading to it. Finally, the unconditional class logits are tallied across all leaf nodes and delivered as the final model output. Note that the illustration in Fig.~\ref{fig:softtree_implement} can be easily generalized to accommodate multiple input features and arbitrary tree structures characterized by the tree depth $d$, which gives rise to $2^{d-1} - 1$ internal nodes and $2^{d-1}$  leaf nodes.

\begin{figure}
    \centering
    \includegraphics[width=6.14in]{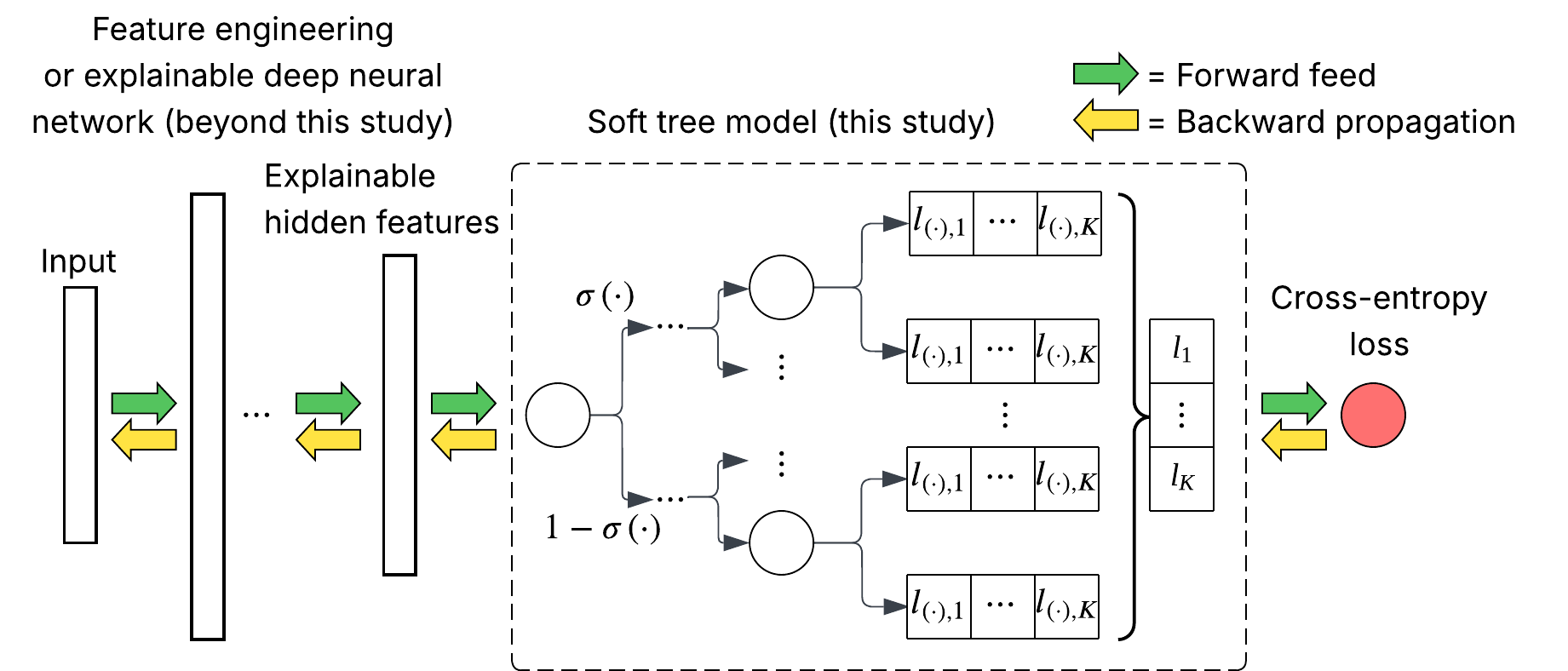}
    \caption{General architecture for supervised classification with soft tree models.}
    \label{fig:softtree_supervised}
\end{figure}

\begin{figure}
    \centering
    \includegraphics[width=6.5in]{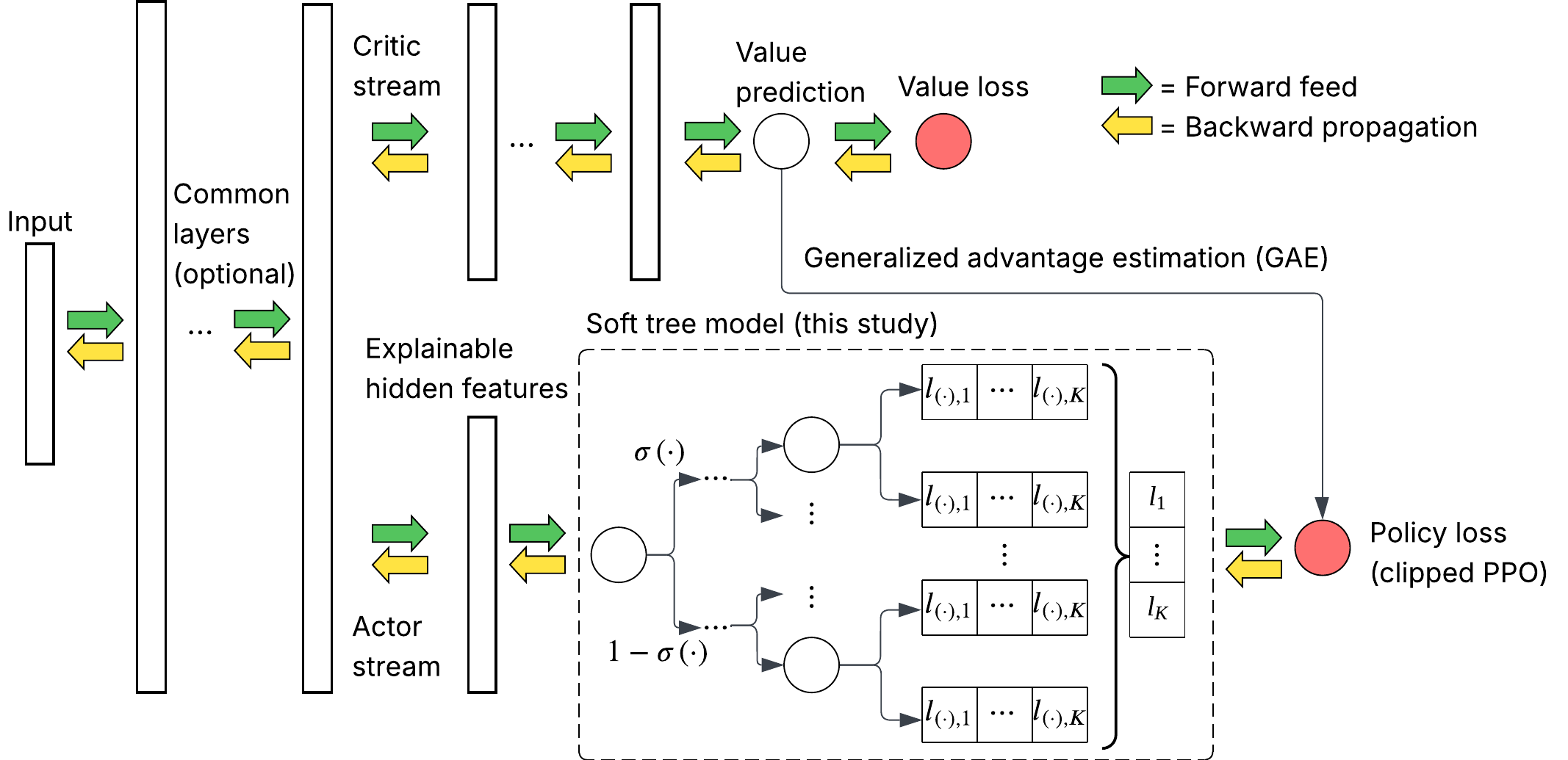}
    \caption{PPO architecture for reinforcement learning with soft tree based actors}
    \label{fig:softtree_rl}
\end{figure}

Once the class (or action) logits are derived, the soft tree model (as implemented in Fig.~\ref{fig:softtree_implement}) can be directly incorporated into both supervised classification and reinforcement learning tasks. In both scenarios, the trainable parameters intrinsic to the soft tree model include the weights and biases of all internal nodes, as well as the conditional logits at all leaf nodes. Hyperparameters associated with the soft tree model include the tree depth, the regularization scaler, and the temperature and temperature annealing parameters governing the sigmoid gating function. Furthermore, these soft tree models can be expanded by stacking other deep learning architectures prior to the linear layer containing the internal nodes. However, to maintain interpretability, separate explainability techniques must be applied to make sense of the hidden input features fed into the linear layer containing the soft tree internal nodes. Fig.~\ref{fig:softtree_supervised} depicts this general architecture for supervised classification tasks, while Fig.~\ref{fig:softtree_rl} presents a RL architecture that employs the proposed soft tree model as the actor within the proximal policy optimization (PPO) algorithm \citep{Schulman2017}. Note that other actor-critic algorithms, e.g. the soft actor-critic algorithm \citep{haarnoja_soft_2018}, can also be adapted by replacing the standard actor network with the proposed soft tree model.

The general architecture and its subsequent conversion to the oblique decision tree form a novel tree-based deep learning framework. Overall, the new approach can be summarized by the following general steps:
\begin{enumerate}
    \item Train a soft tree model with appropriate regularization that can later simplify the tree structure while maintaining satisfactory model performance.
    \item \label{step:anneal} Freeze the trained soft tree into an oblique decision tree. To maintain model performance following conversion, ensure the final soft tree model utilizes a low-temperature sigmoid gating function. Use temperature annealing if necessary.
    \item \label{step:prune} Prune the oblique decision tree based on a prescribed pruning threshold for feature weights and the pruning routines described previously.
    \item Inspect the model interpretability and validate the model performance. Repeat steps 1 through 3 as needed to strike a balance between interpretability and performance.
\end{enumerate}

\section{Validation via Supervised Learning Experiments}
Although the ultimate goal of this study is to optimize life-cycle policies using RL, directly applying the methodology to RL tasks may obscure the benefits and limitations of the proposed soft tree model and its subsequent extraction of oblique decision trees. Because decisions in bridge life-cycle management typically involve discrete options, supervised classification experiments are conducted first. These experiments investigate the fundamental characteristics of the proposed deep learning framework, taking advantage of a simpler loss function, easier training processes, and clearer benchmarks compared to full-fledged RL tasks. Specifically, these numerical experiments are designed to investigate the following characteristics of the new approach:
\begin{itemize}
    \item The expressiveness of the soft tree models in capturing highly nonlinear decision boundaries.
    \item The effects of temperature and temperature annealing schedules.
    \item The tradeoffs between soft tree complexity and model performance.
\end{itemize}

\subsection{Dataset for Classification}
\begin{figure}
    \centering
    \includegraphics[width=3.25in]{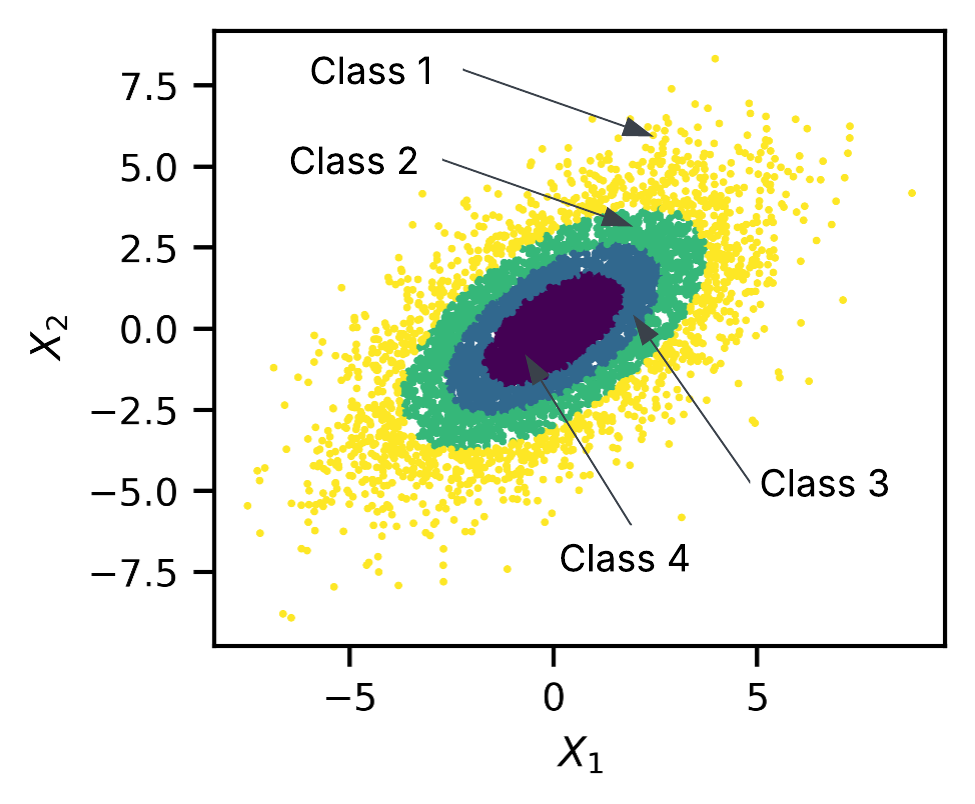}
    \caption{Labeled synthetic data for supervised classification experiments}
    \label{fig:class_data}
\end{figure}

A simple synthetic dataset is leveraged for this investigation. Each data point comprises two real-number features and is assigned to one of four classes. Following a strategy similar to \cite{zhu_multi-class_2009}, the synthetic data are generated using bivariate Gaussian distribution and labeled based on quartiles. A correlation between the two features is introduced to yield nonlinear and oblique decision boundaries. Specifically, samples of input data are generated by two random variables ($X_1$ and $X_2$) defined as follows:
\begin{equation}
    \begin{split}
        X_1 &= 2 Z_1 - Z_2 \\
        X_2 &= 2 Z_1 + Z_2
    \end{split}
    \label{eq:synthetic_data}
\end{equation}
where $Z_1$ and $Z_2$ = two independent standard normal random variables. The bivariate cumulative distribution function (CDF) value associated with a sample of $(z_1, z_2)$ is used to label the corresponding input sample $(x_1, x_2)$. As a result, the two input features have a mean value of 0, a variance of 5, and correlation coefficient equal to 0.6. Fig.~\ref{fig:class_data} shows 10,000 labeled synthetic data points used in the following investigation.

\subsection{Capability of Soft Tree Models}
Using the database described above, four different classification models are trained, including two directly interpretable models (i.e., standard and oblique decision trees), a conventional fully connected neural network, and the proposed soft tree model. The entire dataset is partitioned into a training set (60\%), a validation set (20\%), and a test set (20\%). Given the abundance of data and the small number of input features, holdout validation (instead of cross validation) is implemented if hyperparameter tuning is needed, i.e., for the neural network model. The predictive performance of the four models is evaluated and compared based on their test set accuracy.

The standard and oblique decision trees are trained using the classification and regression trees (CART) algorithm \citep{Geron2019} and the oblique classifier (OC) algorithm \citep{murthy_system_1994}, respectively. It is important to note that both algorithms are non-parametric methods not compatible with gradient-based deep learning algorithms. They are implemented herein to demonstrate the limited capability of conventional tree models in representing nonlinear decision boundaries, essentially highlighting the price of interpretability. As a result, the maximum depth for both models is set to 5, resulting in a maximum of 15 internal nodes and 16 leaf nodes.

For the conventional neural network, two hidden layers each with 18 neurons are used, resulting in a total of 472 trainable parameters. This specific neural network layout is selected because (a) further increasing the number of neurons brings in only a marginal increase in the validation accuracy (less than 5\%), and (b) the layout allows for the creation of a soft tree model with a comparable number of trainable parameters. For the neural network training, the Adam optimizer \citep{Geron2019} is used with a learning rate of 0.002. The training is conducted for 100 iterations with a batch size of 32 data points per iteration.

Finally, a soft tree model with a maximum depth of seven is created and trained according to the general architecture depicted in Fig.~\ref{fig:softtree_supervised}. To facilitate interpretation and given the small number of input features, the input variables are directly fed into the soft tree model without any feature engineering or the addition of any preprocessing layers. This soft tree model possesses 63 internal nodes and 64 leaf nodes, resulting in a total of 445 trainable parameters, a number only slightly smaller than (but still comparable to) that of the neural network model. The same optimizer, batch size, and number of iterations are adopted as those used for the neural network. To focus on the investigation of model capability, no regularization penalty is applied, and a fixed temperature at $T=1$ is used without any temperature annealing for the sigmoid gating function.

Table~\ref{table:softtree_capability} presents the model accuracy scores across all four models. It can be seen that although the standard decision tree offers the strongest interpretability, its reliance on axis-aligned splits results in the lowest performance. An oblique decision tree of the same depth can slightly improve the model performance by engaging both input features at each internal node. Nonetheless, similar to the standard tree, the simple structure of the oblique tree, while ensuring interpretability, greatly limits the overall predictive capability. By contrast, the neural network and the soft tree models significantly outperform the standard and oblique decision trees, albeit with the trade-offs of requiring over 400 model parameters and offering little to none model interpretability.

\begin{table}
    \centering
    \caption{Model performance of different classification models.}
    \small
    \begin{tabular}{l l l l}
            \hline
            Model & Training accuracy (\%) & Validation accuracy (\%) & Test accuracy (\%) \\ \hline
            Standard tree  & 57.78 & 55.90 & 57.10 \\
            Oblique tree   & 60.52 & 58.90 & 59.75 \\
            Neural network & 98.50 & 98.25 & 97.80 \\
            Soft tree      & 90.53 & 90.35 & 90.35 \\ \hline
    \end{tabular}
    \normalsize
    \label{table:softtree_capability}
\end{table}

Between the neural network and the soft tree, the model performance is fairly similar, with the soft tree model performing slightly worse than the neural network. This slight performance disparity can be attributed to the fact that the soft tree model is built upon a wide (63 neurons) and shallow (one-layer) linear layer, as explained in Fig.~\ref{fig:softtree_implement}(a). It is commonly believed that a wide and shallow neural network does not perform as well as  a narrow yet deep neural network when given the same number of model parameters \citep{Geron2019}. This insight seems to apply to the soft tree model here as well. In summary, the soft tree model proposed herein is nearly as capable as a neural network but is slightly less parameter-efficient.

\subsection{Effects of Temperature and Temperature Annealing}
The temperature of the sigmoid function influences how closely it resembles a unit step function, which in turn affects the performance discrepancy between the soft tree and the frozen tree, as well as the subsequent oblique tree after pruning. In order to obtain soft tree models with low-temperature gating functions, an annealing procedure is embedded in the training process, following the procedures described in the Methodology section. Table~\ref{table:class_temp} presents the test accuracy scores associated with different temperatures or temperature annealing schedules for the classification experiments considered herein. All soft trees have a maximum depth of seven, and are trained using the same optimizer, learning rate, and batch and iteration setups described previously. Regularization is investigated separately later and is not applied herein.

\begin{table}
    \centering
    \caption{Test accuracy of soft and frozen trees associated with different temperatures and temperature annealing schedules.}
    \small
    \begin{tabular}{l l l l}
        \hline
        Temperature and annealing & Soft tree score (\%) & Frozen tree score (\%) & Relative difference \\
        \hline
        Fixed at 100         & 46.90  & 27.05 & 42.32\% \\
        Fixed at 1           & 90.35  & 83.10 & 8.02\% \\
        Fixed at 0.01        & 77.55  & 76.05 & 1.93\% \\
        Anneal: 100 to 0.01  & 80.70  & 80.65 & 0.06\% \\
        Anneal: 1 to 0.01    & 91.85  & 91.80 & 0.05\% \\
        \hline
        \multicolumn{4}{l}{\textbf{Note}: ``Frozen'' tree refers to the full oblique tree converted from the soft tree baseline prior to pruning.}
    \end{tabular}
    \normalsize
    \label{table:class_temp}
\end{table}

Three fixed temperatures, i.e., $T=100, 1, \text{and } 0.01$, are first implemented. The results in Table~\ref{table:class_temp} confirm that the relative difference between the soft tree and the frozen tree performance decreases along with the drop of temperature. This observation indicates that to preserve the performance of a soft tree model, the temperature should be sufficiently low. On the other hand, the temperature parameter affects the performance of the baseline soft tree model. In particular, when the temperature is high ($T = 100$), the sigmoid gating function is close to a linear function with near-zero gradients, as demonstrated in Fig.~\ref{fig:sigmoid}. This causes inferior model performance compared to models with lower temperatures. At the other extreme, directly using low-temperature sigmoid gating functions does not lead to effective learning either, as suggested Table~\ref{table:class_temp}. This is attributed to the numerical instability of the gradient of a low-temperature sigmoid gating function.

The results associated with the three fixed temperatures further confirms the necessity of applying temperature annealing schedules. Two annealing schedules ending at $T_{\min} = 0.01$ are explored, focusing on the effects of the starting temperature $T_0$. The results in Table~\ref{table:class_temp} indicate that due to the low ending temperature, annealing leads to closely aligned model test scores between the soft and frozen trees, while maintaining satisfactory model performance. Between the two annealing schedules, the starting temperature plays a role in the model performance: an unnecessarily high starting temperature can actually compromise the model performance. In practice, the starting temperature should be carefully selected based on the range of input features. If min-max or standard input scaling \citep{Geron2019} is applied, our experience suggests that a starting temperature of $T_0=1$ should be appropriate.

\subsection{Effects of Regularization and Pruning}
\begin{figure}
    \centering
    \includegraphics[width=6.25in]{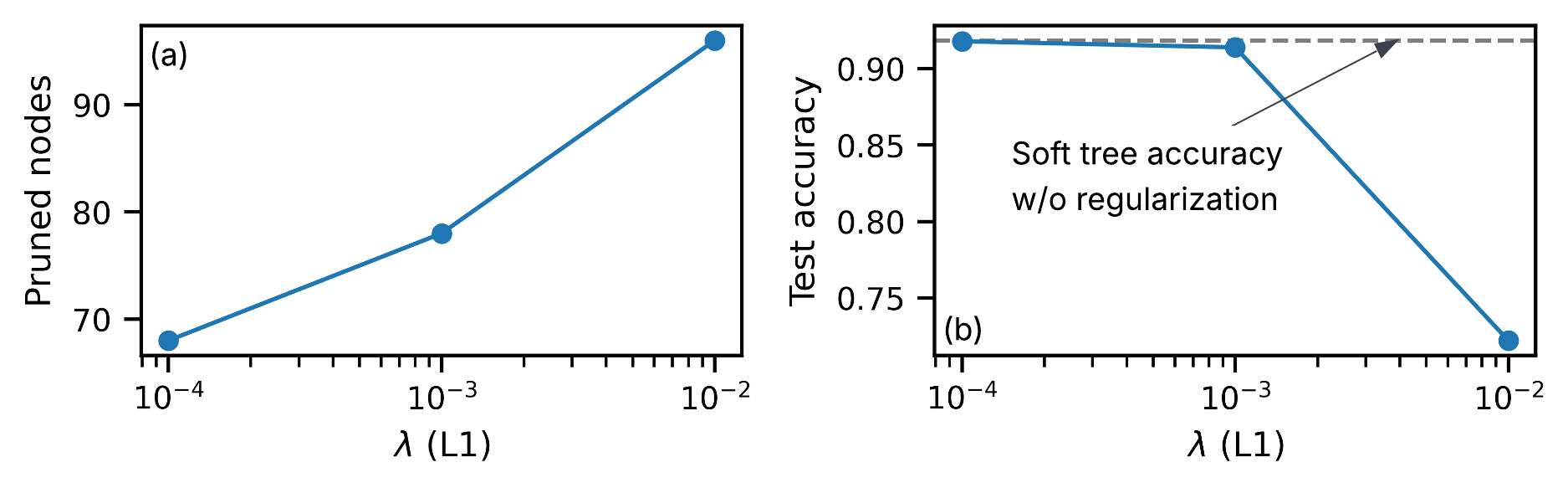}
    \caption{Effects of $L_1$ regularization on (a) number of pruned nodes and (b) model performance}
    \label{fig:regularization}
\end{figure}

Using the annealing schedule from $T=1$ to $T=0.01$, the effects of $L_1$ regularization are investigated using the same classification dataset and the same baseline 7-depth soft tree model. The strength of the regularization is controlled by a hyperparameter $\lambda$, which scales the regularization penalty before adding it to the cross-entropy loss for training. A larger scaler  places more emphasis on regulating model complexity compared to the model performance. In particular, a stronger regularization may push more feature weights below a prescribed threshold, leading to more trivial nodes being pruned from the frozen tree. Fig.~\ref{fig:regularization} presents the effects of $L_1$ regularization across three tested scalers: $10^{-4}$, $10^{-3}$, and $10^{-2}$. Fig.~\ref{fig:regularization} confirms that $L_1$ regularization can effectively prune the frozen tree, with more nodes being pruned as the regularization scaler increases. However, an excessively high regularization scaler (e.g., 0.01 tested herein) compromises the performance of both the baseline soft tree model and the corresponding pruned oblique tree.

\begin{figure}
    \centering
    \includegraphics[width=6.25in]{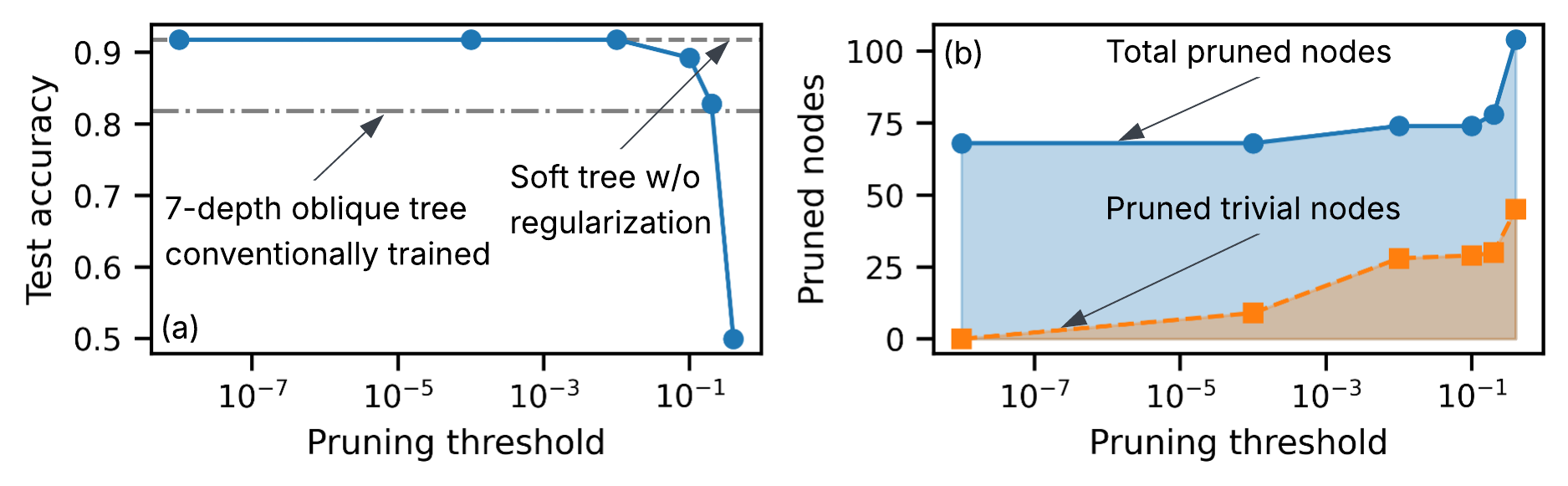}
    \caption{Effects of pruning threshold on (a) model performance and (b) pruned nodes.}
    \label{fig:pruning_threshold}
\end{figure}

The number of the pruned nodes, as summarized in Fig.~\ref{fig:regularization}, is determined based on a pruning threshold of $\epsilon = 10^{-4}$, beneath which a feature weight is forced to zero (i.e., the corresponding feature is turned off). Focusing on the $L_1$ regularization with a scaler of $10^{-4}$, a series of pruning thresholds is evaluated to understand their specific effects. The tested thresholds are $10^{-8}$, $10^{-4}$, $10^{-3}$, $10^{-2}$, $0.1$, $0.2$, and $0.4$. Fig.~\ref{fig:pruning_threshold} illustrates the influence of these pruning thresholds on both model performance (measured again by test accuracy) and the number of pruned nodes. As expected, larger threshold values result in a greater number of pruned nodes. However, even if no nodes are explicitly shut off (e.g., when using the pruning threshold of $10^{-8}$), a substantial number of nodes are still pruned by the routine in Fig.~\ref{fig:prune_infeasible}, because they are along infeasible paths.  Additionally, as long as a reasonable value is used (e.g., below $0.1$), the model performance seems insensitive to the specific pruning threshold chosen, and outperforms the conventional OC algorithm.

\section{Application for Bridge Life-cycle Management}

\subsection{Establishing Training Environment}
Building on the findings from the supervised learning investigation, the same interpretable deep learning approach is applied to RL for element-level life-cycle management. In the context of RL, an environment must first be established to allow the interaction between various life-cycle actions and the ``reward'' associated with the CS distribution of a key bridge element. To this end, an environment is created for steel girder elements, specifically the NBE 107 \citep{AASHTO2019}, defined as a linear foot of a steel girder in steel girder bridges. As a simplification, this study assumes that the reliability of a steel girder bridge is controlled by the load-carrying capacity of its superstructure, which in turn is governed exclusively by the CS distribution of all girders. Additional reliability-critical elements may be included to better align the structural condition with structural reliability. However, establishing such detailed models is beyond the scope of this study.

According to the bridge management convention, the deterioration of a bridge element is modeled herein as a Markov deterioration process with four CSs. The following transition matrix governs the progression from one CS to another:
\begin{equation}
    \mathbf{T}(a = 0) = 
    \begin{bmatrix}
        0.9381 & 0.0619 & 0      & 0      \\
        0      & 0.9356 & 0.0644 & 0      \\
        0      & 0      & 0.8888 & 0.1112 \\
        0      & 0      & 0      & 1
    \end{bmatrix}
    \label{eq:transition_a0}
\end{equation}
where $\mathbf{T}(a=0)$ = transition matrix under a ``do-nothing'' action (i.e., $a=0$). The row index represents the current CS, while the column index denotes the CS transitioned to after a year. Under this convention, for instance, 0.9381 is the annual probability of staying in CS1, while 0.0619 is the annual transition probability from CS1 to CS2. The transition probabilities are derived from \cite{Thompson1998} for the same steel girder element. It should be noted that the original transition matrix in \cite{Thompson1998} uses five CSs associated with the commonly recognized (CoRe) bridge elements, which were defined in the now obsolete pre-2010 AASHTO Guide \citep{aashto_aashto_2002}. Therefore, a conversion is conducted in this study based on the mapping between the CS definitions of a CoRe element and an NBE, as elaborated in the Appendix of the 2010 AASHTO Manual \citep{AASHTO2010}.

During the service life of a bridge, five general life-cycle actions, also known as network-level actions \citep{fhwa_bridge_2020}, are considered. They are referred to as maintenance, repair, rehabilitation, and replacement, listed in the ascending order of their effectiveness in improving bridge condition. The transition matrices given the implementation of these actions are assumed as follows:
\begin{equation}
    \mathbf{T}(a = 1;2;3;4) = 
    \begin{bmatrix}
        0.99; 1; 1; 1      & 0.01;0;0;0         & 0;0;0;0           & 0;0;0;0      \\
        0.15;0.25;0.5;1    & 0.975;0.725;0.5;0  & 0.01;0.025;0;0    & 0;0;0;0      \\
        0;0;0.4;1          & 0.03;0.5;0.5;0     & 0.95;0.45;0.1;0   & 0.02;0.05;0;0 \\
        0;0;0.4;1          & 0;0;0.5;0          & 0;0.5;0.1;0       & 1;0.5;0;0
    \end{bmatrix}
    \label{eq:transition_a1_4}\\  
\end{equation}
where $a=1\dots4$ = maintenance, repair, rehabilitation, and replacement, respectively. The costs per element for these life-cycle actions, including the ``do-nothing'' option, are assumed to be 0 ($a=0$), 10 ($a=1$), 100 ($a=2$), 1,000 ($a=4$), and 2,000 ($a=5$) monetary units, respectively.

To reflect the relationship between the element CS and the structural risk, it is assumed that the CS distribution within a girder bridge can be converted into the failure probability using the following equation:
\begin{equation}
   p_f(\mathbf{s}) = \mathbf{s}^\intercal \Phi \left[ -\boldsymbol{\beta} \right]
        \equiv s_1 \Phi[-\beta_1] + s_2 \Phi[-\beta_2] + s_3 \Phi[-\beta_3] + s_4 \Phi[-\beta_4] 
    \label{eq:pf_s}
\end{equation}
where $s_1 \dots s_4$ = proportions of CS1 through CS4, respectively; $\beta_1 \dots \beta_4$ = reliability indices of a structure if all its elements are in CS1, CS2, CS3, and CS4, respectively; $\Phi(\cdot)$ = CDF of the standard normal distribution. Following common target reliability indices \citep{Liu2021}, this study assumes that $\boldsymbol{\beta}=[4.2, 3.5, 3.0, 2.5]^\intercal$. Consequently, the annual failure risk $R_f$ is expressed as:
\begin{equation}
    R_f (\mathbf{s}) = C_f \cdot p_f(\mathbf{s})
    \label{eq:risk}
\end{equation}
where $C_f$ = failure cost per element, assumed to be 100,000 monetary units per element. Note that since both actions and failure costs are defined per element, the number of elements can be omitted from the cost and risk definitions. The life-cycle cost thus obtained represents the long-term cumulative cost per element.

\begin{figure}
    \centering
    \includegraphics[width=5.625in]{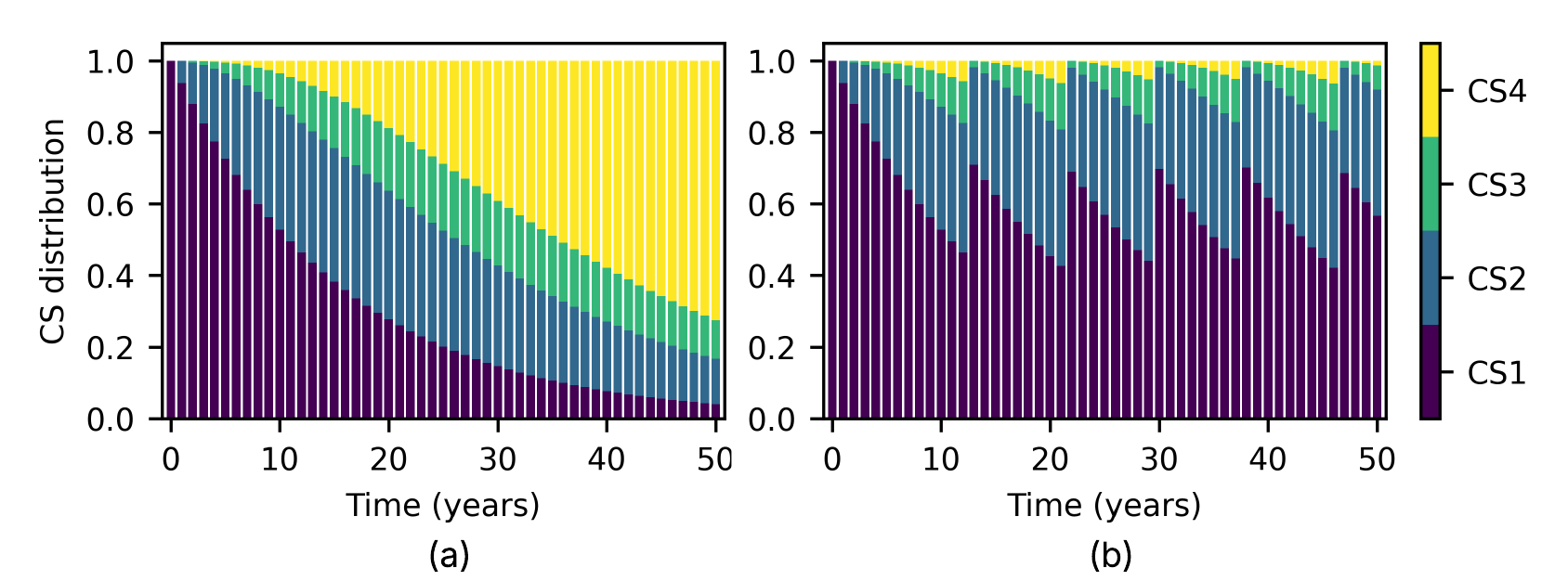}
    \caption{CS evolution under two policies: (a) do-nothing; (b) conduct rehabilitation ($a=3$) if CS4 percentage is greater than 5\%, do nothing otherwise.}
    \label{fig:cs_evolve}
\end{figure}

At each time step (typically one year in bridge management), life-cycle actions are decided based on the bridge CS vector $\mathbf{s}$. Given a selected action, the bridge CS distribution in the next time step can be expressed as
\begin{equation}
    \mathbf{s} (t+1) = \mathbf{T}(a)^\intercal \mathbf{s} (t)
    \label{eq:transition}
\end{equation}
where $\mathbf{s} (t)$ and $\mathbf{s} (t+1)$ = bridge CS vector at time step $t$ and the immediate next step $t+1$, respectively. To formalize this decision-making process, a life-cycle policy can be represented by a parameterized policy function denoted as follows:
\begin{equation}
    \pi \left(t_r,\mathbf{s}; \boldsymbol{\theta} \right)
    \label{eq:policy}
\end{equation}
where $t_r$ = remaining service life or remaining time steps until the end of the decision horizon; $\boldsymbol{\theta}$ = function parameters to be optimized using RL. For long decision horizons, a policy becomes independent to $t_r$, giving rise to a so-called stationary policy that is exclusively linked to the bridge CS vector. Fig.~\ref{fig:cs_evolve} shows the CS evolution over a 50-year period under two hypothetical life-cycle policies, simulated using the developed environment. 

It should be noted that the formulation above defines the bridge condition using the bridge CS vector $\mathbf{s}$. This representation is fundamentally different from previously developed environments, where it is typically assumed that the bridge condition is represented by a single CS or GCR \citep{yang_deep_2022}. According to this vector-based state definition, a new bridge has its system state represented as $\mathbf{s}=[1,0,0,0]^\intercal$. Given a life-cycle policy, the evaluation of the bridge state is uniquely determined, as illustrated in Fig.~\ref{fig:cs_evolve}. As a result, if a policy is trained using RL starting from a new bridge only, much of the state space remains unexplored. In practice, the state space of the actual bridge condition is much more diverse due to (a) the limited numbers of elements in a bridge and (b) the imperfections of Markov assumptions. Hence, a policy trained exclusively from a new bridge state may fail to effectively reduce the life-cycle cost of an existing bridge.

To address this issue, the environment is configured to restart from a randomized bridge condition drawn from a joint distribution of different CSs. Given that a CS vector consists of four proportions that sum to one, a Dirichlet distribution is employed to generate the random starting state for each training episode. The parameters of the Dirichlet distribution are estimated based on the CS data obtained from the FHWA InfoBridge database \citep{fhwa_fhwa_2024} for steel girder bridges owned by the Oregon Department of Transportation (ODOT) as of December 2025. Specifically, 222 samples of CS vectors are extracted from the NBE 107 data across 222 ODOT-owned bridges. Maximum likelihood estimation is then conducted to determine the theta vector defined for the Dirichlet distribution \citep{kotz_chapter_2000}. Table~\ref{table:dirichlet_fitting} compares the mean and standard deviation (StD) of each CS proportion, obtained from the InfoBridge data and the fitted Dirichlet distribution, respectively. It is important to note that the fitted Dirichlet distribution provides higher proportions for CS3 and CS4 compared to the InfoBridge data. This indicates that employing a random restart with the Dirichlet distribution can cover a broader initial state space than what the current ODOT bridge stock implies. This over-coverage is highly desirable because it can account for future deterioration that is not yet reflected in the current CS data. This serves as an additional reason why the Dirichlet distribution is used instead of simply resampling from the current ODOT stock.

\begin{table}
    \centering
    \caption{Mean and StD of CS proportions}
    \small
    \begin{tabular}{lllll}
        \hline
        CS & \multicolumn{2}{l}{InfoBridge Data} & \multicolumn{2}{l}{Fitted Dirichlet} \\ 
        \cline{2-5}
           & Mean (\%) & StD (\%) & Mean (\%) & StD (\%) \\ 
        \hline
        CS1 & 63.5 & 37.5 & 42.7 & 42.6 \\
        CS2 & 33.3 & 35.4 & 31.8 & 40.1 \\
        CS3 & 3.2 & 8.9 & 14.3 & 30.1 \\
        CS4 & $\sim$0 & $\sim$0 & 11.2 & 27.2 \\ 
        \hline
        \multicolumn{5}{p{0.5\textwidth}}{
            \textbf{Note}: Fitted Dirichlet is parameterized by theta vector of $[0.1496, 0.1114, 0.0500, 0.0393]^\intercal$
        }
    \end{tabular}
    \normalsize
    \label{table:dirichlet_fitting}
\end{table}

\subsection{Optimizing Life-cycle Policy with RL}
For existing bridges, the primary goal of risk-based life-cycle management is to minimize the prospective life-cycle cost, which ignores sunk cost and encompasses the cost of all life-cycle actions in the remaining service life and the deterioration-induced failure risks accumulated annually over that same period. This process of life-cycle optimization can be expressed by the following equation:
\begin{equation}
    \boldsymbol{\theta}^* = \arg\min_{\boldsymbol{\theta}} {\left[
        \sum_{t=1}^H {
            \gamma^t R_f\left( \mathbf{s}(t)\right) +
            \gamma^t C_a \left( \pi\left( \mathbf{s}(t); \boldsymbol{\theta} \right) \right)
        }
    \right]}
    \label{eq:lifecycle_opt}
\end{equation}
where $\boldsymbol{\theta}^*$ = optimal parameters for the life-cycle policy; $H$ = decision horizon in years; $\gamma$ = discount factor related to the nominal discount rate; $\gamma=1/1.03$ is assumed in this study, corresponding to an annual discount rate of 3\%. A decision horizon of 200 years is considered herein, aligning with previous studies \citep{yang_deep_2022}. Operating within computational capabilities, this long decision horizon allows for a focus on stationary policies based solely on the bridge CS vector, while simultaneously eliminating the need to estimate salvage values at the end of the decision horizon.

The life-cycle optimization in Eq.~\ref{eq:lifecycle_opt} is carried out using the PPO algorithm coupled with the previously proposed soft tree model and the general architecture presented in Fig.~\ref{fig:softtree_rl}. To preserve interpretability, the preprocessing layers illustrated in Fig.~\ref{fig:softtree_rl}, as well as the shared common layers between the actor and critic modules, are omitted; instead, the CS vector is fed directly into the actor model.

To compare the soft tree model with non-interpretable models, the actor in the PPO algorithm is implemented using both the conventional neural network and the proposed soft tree model. Table~\ref{table:actor_hyperparams} summarizes the hyperparameters associated with both types of actors. The total number of trainable parameters for the neural network actor is 4,745, while an 11-depth soft tree actor has 9,216 trainable parameters (approximately double that of the neural network). This deliberate increase in trainable parameters for the soft tree actor compensates for (a) the slightly inferior model performance observed in the supervised learning investigation, and (b) the potential elimination of internal nodes and features when the soft tree model is later converted into the oblique decision tree. Apart from the actor hyperparameters, Table~\ref{table:ppo_hyperparams} presents the common hyperparameters related to the critic model and the PPO algorithm. Note that a neural network based critic model is acceptable because only the actor model needs to be interpretable once fully trained.

\begin{table}
    \centering
    \caption{Hyperparameters for different actor models.}
    \small
    \begin{tabular}{l l p{0.5\textwidth}}
        \hline
        Actor model & Hyperparameter & Value \\
        \hline
        Neural network  & Network layout  & Two fully connected hidden layers, each with 64 neurons and an expoential linear unit (ELU) as the activation function\\
        Soft tree  & Maximum depth & 11\\
                   & $L_1$ regularization scaler & $0.01$\\
                   & Initial temperature & 1\\
                   & Temperature update & 0.9550 (from $T=1$ to $T=0.01$ in 100 stages)\\
                   & Update frequency & 1 batch per stage\\
        \hline
    \end{tabular}
    \normalsize
    \label{table:actor_hyperparams}
\end{table}

\begin{table}
    \centering
    \caption{Hyperparameters for critic model and PPO training (common to both types of actors).}
    \small
    \begin{tabular}{l l p{0.5\textwidth}}
        \hline
        Category & Hyperparameter & Value \\
        \hline
        Critic model  & Network layout  & Three fully connected hidden layers, each with 32 neurons and an ELU as the activation function.\\
        PPO loss   & Clipping epsilon  & 0.01\\
                   & Entropy coefficient & 0.05\\
                   & Critic coefficient & 0.5\\
        General advantage & Lambda & 0.95\\
        Training   & Total batches & 100\\
                   & Batch size & 100 episodes (2,000 time steps)\\
                   & Minibatch size & 200 time steps\\
                   & Minibatches per batch & 100\\
        Adam optimizer & learning rate & 0.001\\
        \hline
    \end{tabular}
    \normalsize
    \label{table:ppo_hyperparams}
\end{table}

\begin{figure}
    \centering
    \includegraphics[width=3.25in]{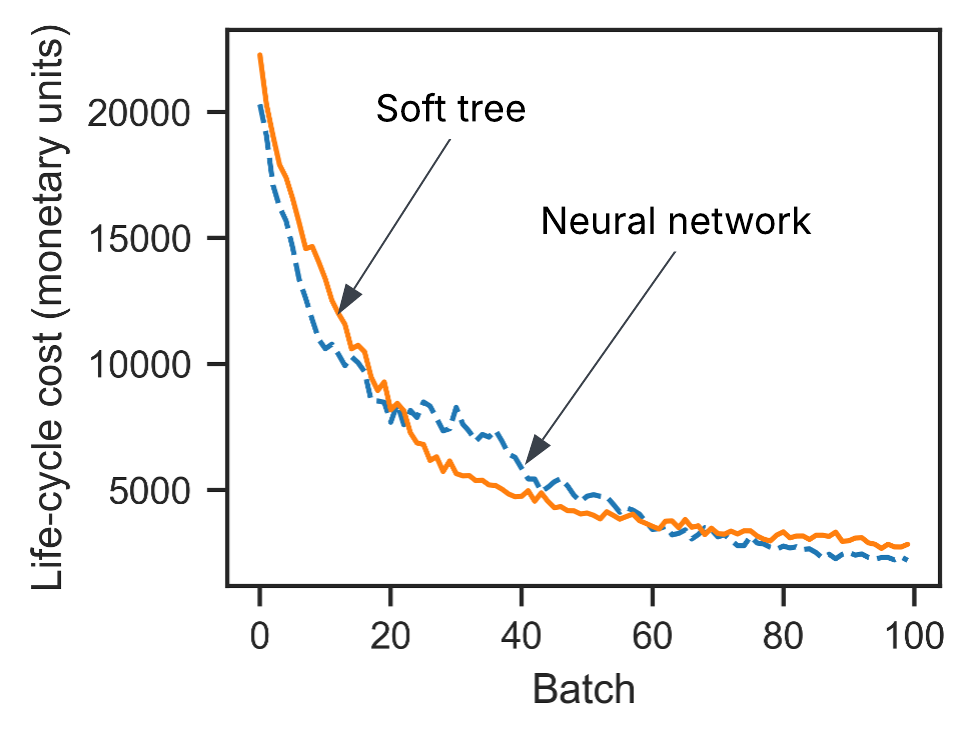}
    \caption{Learning curves associated with both types of actor models.}
    \label{fig:RL_learning}
\end{figure}

Based on the hyperparameters detailed in Tables~\ref{table:actor_hyperparams} and \ref{table:ppo_hyperparams}, Fig.~\ref{fig:RL_learning} presents the learning curves for both the neural network and soft tree actors. As depicted, both actor models can successfully reduce life-cycle cost as training progresses. Ultimately, both models yield similar performance in terms of the life-cycle costs at the end of training. To further compare the policies obtained from both actors, the trained policies are validated using an additional 1,000 episodes. Note that during validation, the actors pick the actions deterministically based on the action with the highest logit. Table~\ref{table:noninterpretable_val} summarizes the average life-cycle costs and their StDs across the 1,000 episodes. Consistent with the supervised learning results, Table~\ref{table:noninterpretable_val} indicates that RL using either a neural network or a soft tree actor can deliver comparable life-cycle costs. It should be noted that these averages reflect the average life-cycle cost per element for a stock of bridges, whose CS distributions can be represented by the fitted Dirichlet distribution. The relatively high variance of the life-cycle costs stems from the fact that the life-cycle cost heavily depends on the initial CS distribution, which has been randomly generated with the Dirichlet distribution. 

\begin{table}
    \centering
    \caption{Validation results (in monetary units) associated with neural network and soft tree actors.}
    \small
    \begin{tabular}{l l l}
        \hline
        Actor model & Average LCC (monetary unit) & StD (monetary unit) \\
        \hline
        Neural network  & 1540.66  & 640.41\\
        Soft tree  & 1577.03 & 696.20\\
        \hline
    \end{tabular}
    \normalsize
    \label{table:noninterpretable_val}
\end{table}

\subsection{Interpretable Life-cycle Policies}
In order to obtain an interpretable policy, two additional regularization scalers have been tested, including $\lambda=0.1 \text{ and } 0.001$. Using 1,000 validation episodes, the average life-cycle costs and StDs associated with these three scalers are summarized in Table~\ref{table:L1_val}. Table~\ref{table:L1_val} confirms that the scaler used previously, $\lambda=0.01$, is sufficiently large to potentially induce substantial pruning, while simultaneously preserving the model performance. Further increasing it to $\lambda=0.1$ drastically degrades the model performance, while reducing it fails to yield any performance gains.

\begin{table}
    \centering
    \caption{Model performance of soft tree policies trained with different $L_1$ regularization scalers.}
    \small
    \begin{tabular}{l l l}
        \hline
        $L_1$ scaler & Average LCC (monetary unit) & StD (monetary unit) \\
        \hline
        0.1 & 3754.33  & 432.58\\
        0.01 (baseline)  & 1577.03 & 696.20\\
        0.001  & 1582.21 & 712.49\\
        \hline
    \end{tabular}
    \normalsize
    \label{table:L1_val}
\end{table}

\begin{figure}
    \centering
    \includegraphics[width=1.621in]{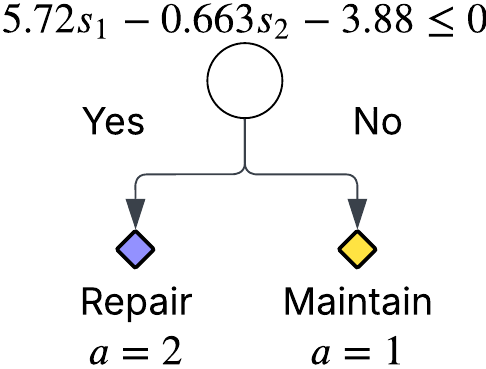}
    \caption{RL-derived oblique decision tree after pruning.}
    \label{fig:pruned_RL}
\end{figure}

From the baseline soft tree model, the corresponding frozen tree is pruned and converted into an oblique decision tree model following routines depicted in Figs.~\ref{fig:prune_trivial} and \ref{fig:prune_infeasible}. To prune trivial nodes, a pruning threshold of $\epsilon=0.001$ is used for weight elimination. In addition, infeasible decision paths are eliminated following the process detailed in Fig.~\ref{fig:prune_infeasible}. After pruning, the simplified oblique decision tree retains only one decision node and two leaf nodes, as illustrated in Fig.~\ref{fig:pruned_RL}. The corresponding life-cycle policy can thus be expressed as the following simple rule:
\begin{equation}
    a = 
    \begin{cases} 
        2 & \text{if} \quad 5.72s_1 - 0.663s_2 - 3.88 \leq 0 \\
        1 & \text{otherwise}
    \end{cases}
    \label{eq:pruned_RL}
\end{equation}

Based on 1,000 validation episodes, the policy represented in Eq.~\ref{eq:pruned_RL} and Fig.~\ref{fig:pruned_RL} yields an average life-cycle cost of 1594.28, with a standard deviation equal to 729.60. Despite its simple structure, this performance remains comparable to that of the complex neural network model and the baseline soft tree model (see Table~\ref{table:noninterpretable_val}). By examining the decision rule, the following insights can be gained from the oblique decision tree:
\begin{enumerate}
    \item Minor life-cycle actions such as maintenance and repair are best implemented annually to reduce long-term life-cycle costs and risks.
    \item The specific decision between maintenance and repair is governed by the proportions of CS1 and CS2. Specifically,
    \begin{enumerate}[label=(\alph*)]
        \item If both are low (e.g., below 0.3), repair is needed;
        \item If CS2 is much more prevalent than CS1, repair is also likely warranted;
        \item Conversely, if CS1 is more prevalent than CS2, only maintenance may be necessary.
    \end{enumerate}
\end{enumerate}

\begin{figure}
    \centering
    \includegraphics[width=2.87in]{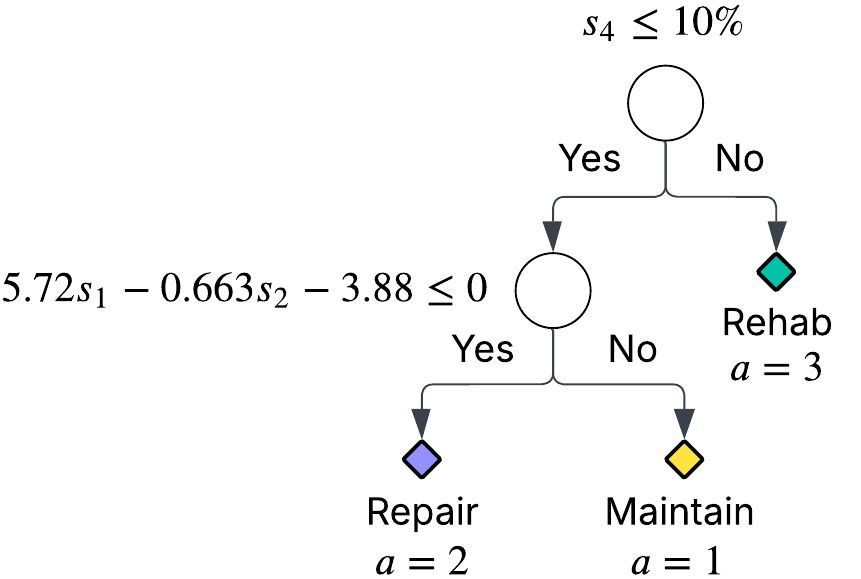}
    \caption{RL-derived oblique decision tree integrated with agency rules.}
    \label{fig:pruned_RL_human}
\end{figure}

This interpretable policy is also conducive to direct integration with existing decision rules or legislative mandates. For instance, if an agency requires a trigger for rehabilitation whenever the proportion of CS4 is greater than 10\%, it can seamlessly add this trigger to the oblique decision tree, resulting in an augmented oblique decision tree as shown in Fig.~\ref{fig:pruned_RL_human}.

\subsection{Comparison with Conventional Methods}
To further benchmark the model performance, the life-cycle policies shown in Figures~\ref{fig:pruned_RL} and \ref{fig:pruned_RL_human} are validated using 1,000 new validation episodes and compared against two conventional optimization methods capable of yielding similarly interpretable life-cycle policies. The first method employs dynamic programming (DP) and assumes that the bridge condition is represented by a single CS rather than the full CS vector. Under this approach, the condition-based action is derived via the value iteration algorithm \citep{bellman_dynamic_1957}, utilizing the same system dynamics established in the training environment. To use this policy, the bridge CS vector is first converted into a single CS based on the most prevalent CS, i.e., the CS with the greatest value in the CS vector. The action corresponding to that CS is then applied. Mathematically, this life-cycle policy can be expressed as follows:
\begin{equation}
    a = 
    \begin{cases} 
        1 & \text{if} \quad \max{(\mathbf{s}) = s_1} \\
        2 & \text{if} \quad \max{(\mathbf{s}) = s_2} \\
        2 & \text{if} \quad \max{(\mathbf{s}) = s_3} \\
        3 & \text{if} \quad \max{(\mathbf{s}) = s_4} 
    \end{cases}
    \label{eq:DP_policy}
\end{equation}

The second method is a reliability-based optimization carried out using the genetic algorithm (GA) \citep{Yang2020} and implemented via the PyGAD package \citep{gad_pygad_2023}. In particular, four reliability index thresholds are treated as decision variables to minimize the life-cycle cost. To use this policy, the CS vector is first converted to the reliability index corresponding to the failure probability in Eq.~\ref{eq:pf_s}. This reliability index is then compared against the optimized reliability thresholds to select the best action out of the five alternatives. Mathematically, the GA-based optimal policy can be expressed as follows:
\begin{equation}
    a = 
    \begin{cases} 
        1 & \text{if} \quad 3.558 < \beta(\mathbf{s}) \leq 4.2 \\
        2 & \text{if} \quad 3.367 < \beta(\mathbf{s}) \leq 3.558 \\
        3 & \text{if} \quad 3.191 < \beta(\mathbf{s}) \leq 3.367 \\
        4 & \text{if} \quad 2.5 \leq \beta(\mathbf{s}) \leq 3.191 \\
    \end{cases}
    \label{eq:GA_policy}
\end{equation}
It is worth mentioning that Eq.~\ref{eq:GA_policy} includes only three reliability index thresholds, because the highest threshold is so close to 4.2 that the ``do-nothing'' option ($a=0$) cannot be realized.

To evaluate their relative effectiveness, these interpretable life-cycle policies, specifically, the oblique decision trees depicted in Figs.~\ref{fig:pruned_RL} and \ref{fig:pruned_RL_human}, the condition-based policy derived via DP (Eq.~\ref{eq:DP_policy}), and the reliability-based policy optimized via GA (Eq.~\ref{eq:GA_policy}), are compared based on their corresponding life-cycle costs. Table~\ref{table:interpretale_val} summarizes the comparison results across 1,000 validation episodes. The results demonstrate that the interpretable oblique decision trees, obtained using the proposed approach, outperforms the conventional life-cycle policies obtained from DP and GA. Interestingly, the augmented oblique tree incorporating the \textit{ad hoc} action trigger performs slightly better than the policy generated solely from RL.

\begin{table}
    \centering
    \caption{Model performance with interpretable life-cycle policies.}
    \small
    \begin{tabular}{l l l}
        \hline
        Policy & Average LCC (monetary unit) & StD (monetary unit) \\
        \hline
        Oblique decision tree (RL) & 1590.86  & 740.31\\
        Oblique decision tree (RL with \textit{ad hoc} rule)  & 1560.96 & 672.09\\
        Condition-based policy (DP)  & 2133.42 & 1178.30\\
        Reliability-based policy (GA)  & 1758.91 & 918.04\\
        \hline
    \end{tabular}
    \normalsize
    \label{table:interpretale_val}
\end{table}

\section{Concluding Remarks}
This study introduces a novel, interpretable deep reinforcement learning framework for bridge life-cycle optimization based on element-level condition data. The proposed methodology utilizes differentiable soft tree models as actor functions, integrated with regularization, pruning rules, and a temperature annealing schedule to extract deterministic and interpretable oblique decision trees that serve as life-cycle policies. Validation through supervised learning experiments demonstrated that these  soft tree models achieve predictive performance comparable to deep neural networks while maintaining intrinsic interpretability. Furthermore, the level of interpretability can be actively controlled by adjusting the strength of regularization and the weight threshold for pruning. In the reinforcement learning application for steel girder elements, the resulting interpretable policy successfully minimized long-range life-cycle costs, outperforming conventional dynamic programming and genetic algorithm-based methods. These results highlight the framework’s capability of providing near-optimal, auditable life-cycle policies that can be readily implemented into existing bridge management systems.

In the proposed approach, the creation of oblique decision tree models, as well as subsequent pruning, relies extensively on recursive algorithms. Although such algorithms yield concise code, they may not be the most computationally efficient implementation. Future studies should focus on developing non-recursive alternatives to improve computational efficiency.

It is also important to note that the specific life-cycle policies obtained from the new framework and presented in this paper are inherently tied to the specific deterioration model, action effects, and the cost assumptions built into the training environment. Therefore, these policies may not be suitable for direct deployment in bridge management unless the underlying assumptions have been thoroughly vetted by bridge owners. Future studies are needed to establish more realistic training environments validated by real-world data and inspection records.

\section{Data Availability Statement}
Some or all data, models, or code generated or used during the study are available in GitHub repositories listed below in accordance with funder data retention policies:
\begin{itemize}
    \item Supervised learning data, as well as code for  data generation, training and validation of soft tree models, freezing soft trees, and pruning frozen trees can be found at \url{https://github.com/InfraRiskGroup/softtree.git}
    \item Code related to the training environment for the reinforcement learning application can be found at \url{https://github.com/InfraRiskGroup/bridge-gym.git}
    \item Code for the training and validation of soft tree actors, and code for extracting oblique decision tree based life-cycle policies can be found at \url{https://github.com/InfraRiskGroup/softtree-RL.git}
\end{itemize}

Some or all data used for generating the tables and figures in the paper are available from the corresponding author by request.

\section{Acknowledgments}
The authors are grateful for the financial support received from the Federal Highway Administration under Contract 693JJ321C000030 and from Portland State University. They would also like to acknowledge the support received from PacTrans, the Region 10 University Transportation Center. The opinions and conclusions presented in this paper are those of the authors and do not necessarily reflect the views of the sponsoring organizations.

\section{Author Contributions}
Seyyed Amirhossein Moayyedi: Data curation; Formal analysis; Investigation; Software; Validation; Writing -- original draft; Writing -- review and editing; David Y. Yang: Conceptualization; Data curation; Formal analysis; Funding acquisition; Investigation; Methodology; Project administration; Software; Supervision; Validation; Visualization; Writing -- original draft; Writing -- review and editing.
%
% Here's the list of references:
%
\bibliographystyle{unsrtnat}
\bibliography{references}

\end{document}